\documentclass[conference]{IEEEtran}
\IEEEoverridecommandlockouts
\usepackage{cite}
\usepackage{amsmath,amssymb,amsfonts}
\usepackage{algorithmic}
\usepackage{subfigure}
\usepackage{graphicx}
\usepackage{textcomp}
\usepackage{makecell}
\usepackage{xcolor}
\usepackage{hyperref}
\usepackage{booktabs}
\usepackage{multirow}
\newcommand{\tabincell}[2]{\begin{tabular}{@{}#1@{}}#2\end{tabular}}
\def\BibTeX{{\rm B\kern-.05em{\sc i\kern-.025em b}\kern-.08em
    T\kern-.1667em\lower.7ex\hbox{E}\kern-.125emX}}
\begin{document}

\title{RACA: Relation-Aware Credit Assignment for Ad-Hoc Cooperation in Multi-Agent Deep Reinforcement Learning}


\author{
\IEEEauthorblockN{Hao Chen${^{1,2}}$, Guangkai Yang${^{1,2}}$, Junge Zhang${^{1,2}}$\IEEEauthorrefmark{2}, Qiyue Yin${^{1,2}}$, Kaiqi Huang${^{1,2,3}}$}
\IEEEauthorblockA{
\textit{$^1$School of Artificial Intelligence, University of Chinese Academy of Sciences}}
\textit{$^2$CRISE, Institute of Automation, Chinese Academy of Sciences}\\
\textit{$^3$CAS Center for Excellence in Brain Science and Intelligence Technology}\\
Beijing 100049, P.R.China\\
\{chenhao2019, yangguangkai2019\}@ia.ac.cn, \{jgzhang, qyyin, kqhuang\}@nlpr.ia.ac.cn
\\
\thanks{\IEEEauthorrefmark{2}Corresponding author.}
}

\maketitle

\begin{abstract}
In recent years, reinforcement learning has faced several challenges in the multi-agent domain, such as the credit assignment issue. Value function factorization emerges as a promising way to handle the credit assignment issue under the centralized training with decentralized execution (CTDE) paradigm. However, existing value function factorization methods cannot deal with ad-hoc cooperation, that is, adapting to new configurations of teammates at test time. Specifically, these methods do not explicitly utilize the relationship between agents and cannot adapt to different sizes of inputs. To address these limitations, we propose a novel method, called Relation-Aware Credit Assignment (RACA), which achieves zero-shot generalization in ad-hoc cooperation scenarios. RACA takes advantage of a graph-based relation encoder to encode the topological structure between agents. Furthermore, RACA utilizes an attention-based observation abstraction mechanism that can generalize to an arbitrary number of teammates with a fixed number of parameters. Experiments demonstrate that our method outperforms baseline methods on the StarCraftII micromanagement benchmark and ad-hoc cooperation scenarios.

\end{abstract}


\section{Introduction}
Cooperative multi-agent reinforcement learning (MARL) plays a vital role in the development of artificial intelligence. Many complex real-world problems are inherently multi-agent systems (MAS), for example, a team of autonomous cars\cite{bhalla2020deep}, sensor networks\cite{ye2015multi}, and power distribution networks\cite{gao2021consensus}. However, multi-agent reinforcement learning has encountered many unique challenges that reinforcement learning has not, such as credit assignment, the non-stationarity of an environment and ad-hoc cooperation.

To effectively deal with these problems, people have witnessed great progress in MARL methods\cite{yang2020multi, foerster2018counterfactual, wang2020roma} in recent years. Among these methods, the centralized training with decentralized execution(CTDE) \cite{Lowe2017MultiAgentAF} paradigm has been widely used to deal with the non-stationarity of the environment. Under this paradigm, value function factorization methods such as QMIX\cite{rashid2018qmix} have shown state-of-the-art performance on challenging tasks such as the StarCraftII micromanagement benchmark SMAC\cite{samvelyan19smac}. In this line of work, each agent has a decentralized Q function and these local Q values are integrated into a global Q value by a mixing network. 

However, in many real-world scenarios, agents are required to be capable of adapting to new configurations of teammates at test time, such as different numbers, categories, and positions of teammates. This is known as ad-hoc cooperation\cite{stone2010ad}. Existing value function factorization methods\cite{rashid2018qmix, Son2019QTRANLT, wang2020qplex, xu2021mmd} lack the ability to deal with ad-hoc cooperation tasks because the global state information used by these methods doesn't explicitly and effectively utilize the relationship between agents. Utilizing the relationship between agents is vital to multi-agent cooperation\cite{liu2020multi} and especially important to ad-hoc cooperation because the relationship is constantly changing during training and testing. Also, these methods cannot adapt to different sizes of inputs during test time. As a result, these methods are trained and tested on the same environment, i.e. the state transition function of the environment remains unchanged, resulting in the inability to cooperate with different configurations of teammates at test time.

In this paper, we address these limitations in ad-hoc cooperation by proposing a novel multi-agent reinforcement learning algorithm called Relation-Aware Credit Assignment (RACA). Our key observation is that, when collaborating with different agents, the relationship between agents is different. Thus, the relationship between agents needs to be fully explored to achieve better ad-hoc cooperation. When we take this relationship between agents, i.e. the underlying topological structure between agents, into consideration, graph-based structure seems a natural choice for its amazing ability to deal with data in an irregular or non-Euclidean domain. Previous works in MARL such as DGN\cite{jiang2018graph} have used graph convolutional network (GCN) to model the communication network between agents. However, they require agents to be able to communicate during testing and cannot fully explore the underlying topological structure between agents.

In this work, we build an undirected graph, where each agent has a corresponding node in the graph. Based on this graph, we use a relation encoder to make use of the topological structure between agents and an attention-based observation abstraction mechanism with a fixed number of parameters to deal with the varying number of teammates. Therefore, RACA can achieve zero-shot generalization in ad-hoc cooperation scenarios without retraining or fine-tuning. We test our method on StarCraft II micromanagement environments and ad-hoc cooperation scenarios. Results show that our method achieves better performance than baseline methods.

In this paper, our contributions can be summarised as:
\begin{itemize}
\item We propose RACA, which makes use of the topological structure between agents during centralized training process to achieve zero-shot generalization in ad-hoc cooperation scenarios.

\item We use a graph-based relation encoder to explore the topological structure between agents. Besides, we use an attention-based observation abstraction mechanism that can generalize to an arbitrary number of teammates with a fixed number of parameters to deal with the varying number of teammates during training and testing.

\item We prove that the proposed algorithm has better performance than the baselines on the StarCraftII micromanagement benchmark SMAC and ad-hoc cooperation scenarios.

\end{itemize}

\section{Background}

\subsection{Dec-POMDP}
In this work, we consider a fully cooperative multi-agent task, which can be described as a decentralized partially observable Markov decision process (Dec-POMDP)\cite{Oliehoek2016ACI} consisting of a tuple $G=\left\langle S,U,P,r,Z,O,n,\gamma\right\rangle$. $s \in S$ is the global state of the environment. At each time step, each agent $a \in A \equiv \{1,...,n\}$ chooses an action $u^a\in U$. The joint action takes the form of $\mathbf{u}\in\mathbf{U}\equiv U^n$. $P(s'|s,\mathbf{u}):S\times\mathbf{U}\times S\rightarrow [0,1]$ represents the state transition function. The same reward function $r(s,\mathbf{u}):S\times\mathbf{U}\rightarrow\mathbb{R}$ is shared among all agents. $\gamma\in[0,1)$ is the discount factor. In the partially observable scenario, the individual observations $z\in Z$ of each agent are obtained according to the observation function $O(s,a):S\times A\rightarrow Z$. Each agent $a$ has its local action-observation history $\tau^a\in T\equiv(Z\times U)^*$. The policy of each agent $a$ takes the form of $\pi^a(u^a|\tau^a):T\times U\rightarrow [0,1]$. The joint policy $\pi$ of all agents has a joint action-value function: $Q^\pi(s_t, \mathbf{u}_t)=\mathbb{E}_{s_{t+1:\infty},\mathbf{u}_{t+1:\infty}} \left[R_t|s_t,\mathbf{u}_t\right]$. Maximizing the discounted return $R_t=\sum^{\infty}_{i=0}\gamma^ir_{t+i}$ is the goal of the Dec-POMDP scenario.

\subsection{Value Function Factorization}
In the cooperative multi-agent reinforcement learning problem, how to assign accurate credit to each agent is a critical challenge known as the credit assignment issue. Value function factorization emerges as a promising way to handle this issue under the CTDE paradigm. The value function factorization methods are based on the Individual-Global-Max (IGM)\cite{Son2019QTRANLT} assumption that the optimal joint actions across agents are consistent with the collection of individual optimal actions of each agent. The IGM assumption can be expressed as:
\begin{equation*}
    \arg \max _{\boldsymbol{u}} Q_{\mathrm{tot}}(\boldsymbol{\tau}, \boldsymbol{u})=\left(\begin{array}{c}
\arg \max _{u^{1}} Q_{1}\left(\tau^{1}, u^{1}\right) \\
\vdots \\
\arg \max _{u^{n}} Q_{n}\left(\tau^{n}, u^{n}\right)
\end{array}\right),
\end{equation*}
where $Q_{\text{tot}}$ and $Q_a$ respectively represent the joint action-value function and the individual action-value function, $\boldsymbol{\tau} \in T^n$ represents the joint action-observation histories of all agents.

VDN\cite{sunehag2018value} attempts to factorize the joint action-value function $Q_{\text{tot}}$ assuming additivity, which can be expressed as: 
\begin{equation*}
    Q_{\text{tot}}(\boldsymbol{\tau}, \boldsymbol{u})=\sum_{a=1}^{n} Q_{a}\left(\tau^{a}, u^{a}\right).
\end{equation*}

QMIX\cite{rashid2018qmix} factorizes the joint action-value function $Q_{\text{tot}}$ by assuming monotonicity, which is guaranteed by restricting the mixing network to have positive weights:
\begin{equation*}
    \frac{\partial Q_{\mathrm{tot}}(\boldsymbol{\tau}, \boldsymbol{u})}{\partial Q_{a}\left(\tau^{a}, u^{a}\right)} \geq 0, \quad \forall a \in \{1,\dots,n\}.
\end{equation*}

QTRAN\cite{Son2019QTRANLT} transforms the original joint action-value function into an easily factorizable one and uses it as a soft regularization constraint. WQMIX\cite{rashid2020weighted} introduces a weighted projection into the monotonic value factorization to place more emphasis on better joint actions. However, none of these value function factorization methods generalize well to ad-hoc cooperation scenarios since these methods fail to effectively make use of the relationship between agents.

\subsection{Graph Convolutional Networks}
Many important real-world applications naturally take the form of graphs, such as traffic flow\cite{guo2019attention}, wireless communications\cite{eisen2020optimal}, interfaces between proteins\cite{fout2017protein}, and social networks\cite{kipf2016semi}. As a kind of graph neural network, graph convolutional neural network (GCN) is often used to process structured input data and incorporate neighborhood information. A GCN takes the feature matrix and the adjacency matrix as inputs and outputs a node-level feature matrix. This is in close resemblance with the convolution operation in convolutional neural networks. Various variants of GCN have been proposed in recent years, such as GAT\cite{Velickovic2018GraphAN}, GraphSAGE\cite{hamilton2017inductive}, and JK-Net\cite{Xu2018RepresentationLO}. 

In MARL, GCN-based architectures have been used to promote communication between agents\cite{jiang2018graph, su2020counterfactual, liu2020multi} and improve sample efficiency\cite{liu2020pic}. However, none of these methods use GCN to promote ad-hoc cooperation by encoding the relationship between agents.

\subsection{Ad-Hoc Cooperation in MARL}
Ad-hoc cooperation has been drawing a lot of attention recently\cite{chen2020aateam, zhang2020multi, mahajan2022generalization} for its promising future in pushing forward the frontier of multi-agent cooperation. The study of ad-hoc cooperation in multi-agent settings has a long history dating back to the early 2000s\cite{bowling2005coordination,stone2010ad}. In ad-hoc cooperation, agents need to learn a policy that can adapt to different teammates at test time. Some methods assume teammate behavior was known\cite{agmon2012leading, stone2009leading}, some methods use explicit hard-coded protocols\cite{tambe1997towards, grosz1996collaborative}, some methods use Monte Carlo tree search to find optimal policy\cite{barrett2011empirical}. 

In recent years, deep learning-based methods was developed, such as pre-training\cite{chen2020aateam, ijcai202166}, population-based training\cite{long2020evolutionary}, online policy adaptation\cite{gu2021online} and adversarial training\cite{li2019robust}. Some methods focus on open ad-hoc teamwork\cite{rahman2021towards}. However, these methods either fail to effectively utilize the relationship between agents or assume prior knowledge of teammates can be accessed, which is impossible in the real world or complex scenarios. Instead of using prior knowledge, our proposed method uses a relation-aware credit assignment mechanism to achieve zero-shot generalization in ad-hoc cooperation scenarios.

\begin{figure*}[htp]
    \centering
    \includegraphics[width = 6 in]{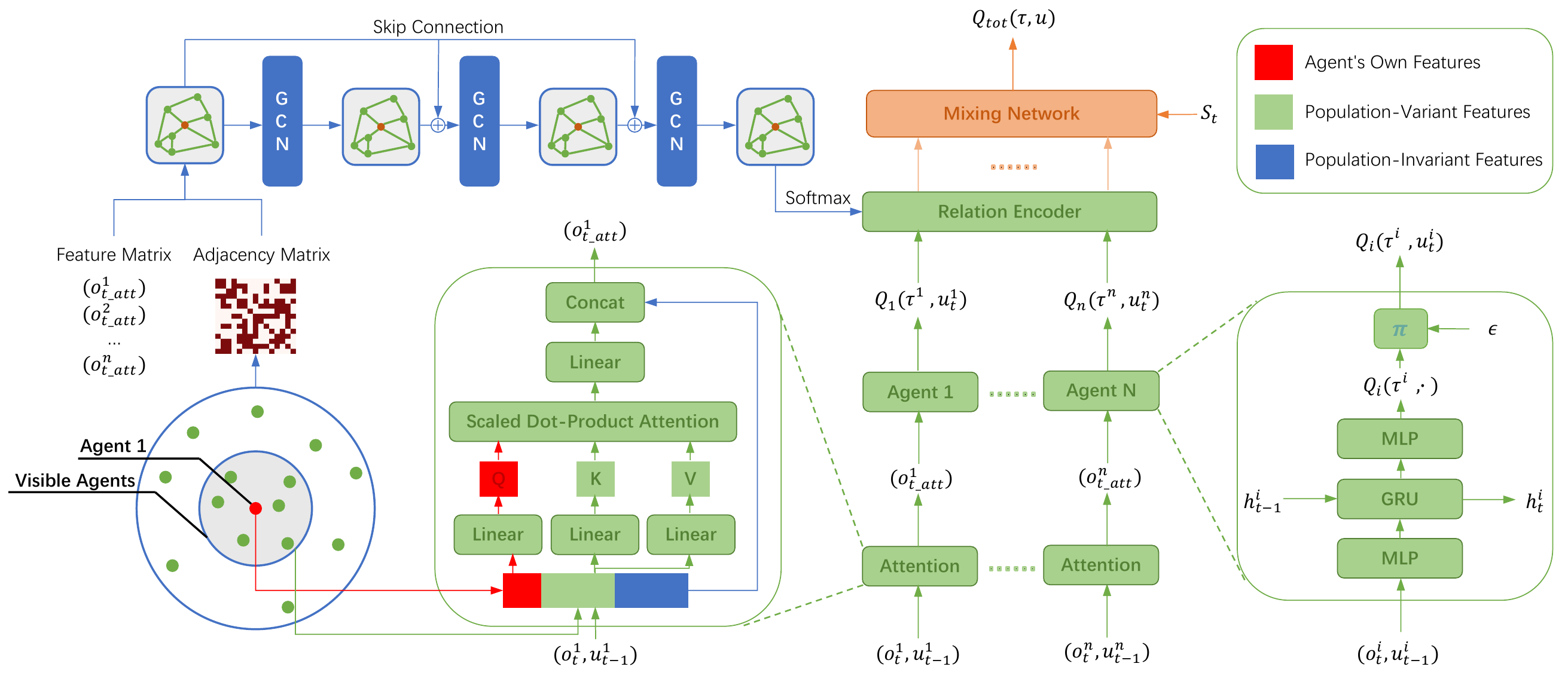}
    \caption{The overall architecture of RACA. The right part is a deep recurrent Q-network, which takes the local action-observation history as input and outputs the individual action-values. The left part is the graph-based relation encoder and the attention-based observation abstraction module.}
    \label{fig:framework}
\end{figure*}

\begin{table*}[ht]
\caption{Maps in different scenarios.}
\label{table:scenario1}
\centering
\begin{tabular}{lccccc}
\hline
Map Name&Ally Units&Enemy Units&Map Type&Critical Challenges&Map Difficulty\\
\hline
& & & &  \\[-6pt]
5m&\tabincell{c}{5 Marines}&\tabincell{c}{5 Marines}&\tabincell{c}{Homogeneous, Symmetric}&Focus Fire&Easy\\
\hline
& & & &  \\[-6pt]
5ma&\tabincell{c}{5 Marauders}&\tabincell{c}{5 Marauders}&\tabincell{c}{Homogeneous, Symmetric}&Focus Fire&Easy\\
\hline
& & & &  \\[-6pt]
1c3s5z&\tabincell{c}{1 Colossus\\3 Stalkers\\5 Zealots}&\tabincell{c}{1 Colossus\\3 Stalkers\\5 Zealots}&\tabincell{c}{Heterogeneous, Symmetric}&\tabincell{c}{Focus Fire, Macro Tactics}&Easy\\
\hline
& & & &  \\[-6pt]
XsYz&\tabincell{c}{X Stalkers\\Y Zealots}&\tabincell{c}{X Stalkers\\Y Zealots}&\tabincell{c}{Heterogeneous, Symmetric}&\tabincell{c}{Focus Fire, Macro Tactics}&Easy\\
\hline
& & & &  \\[-6pt]
Xm\_vs\_Ym&X Marines&Y Marines&\tabincell{c}{Homogeneous, Asymmetric}&Focus Fire&\tabincell{c}{Easy(Y=X), Hard(Y=X+1)} \\
\hline
& & & &  \\[-6pt]
MMM2&\tabincell{c}{1 Medivac\\2 Marauders\\7 Marines}&\tabincell{c}{1 Medivac\\3 Marauder\\8 Marines}&\tabincell{c}{Heterogeneous, Asymmetric}&\tabincell{c}{Focus Fire, Macro Tactics}&Super Hard\\
\hline
\end{tabular}
\end{table*}

\section{Relation-Aware Credit Assignment}
In this chapter, we propose a novel relation-aware credit assignment method called RACA, a new approach to deal with the ad-hoc cooperation problem in multi-agent reinforcement learning. RACA encodes the relationship between agents into the mixing network through a graph-based relation encoder. Furthermore, to deal with the varying number of teammates in ad-hoc cooperation, RACA utilizes an attention-based observation abstraction mechanism. RACA cannot only make full use of the topological structure between agents, but also achieve zero-shot generalization in ad-hoc cooperation scenarios without retraining or fine-tuning.

\subsection{Graph-Based Relation Encoder}
To effectively utilize the relationship between agents, we use a graph-based relation encoder to encode the topological structure between agents. First, we construct the multi-agent environment as a graph $G=(V, E)$, where each agent $i$ in the environment is represented by a node in the graph $v \in V$. Each node $v$ has a set of neighbors $\mathcal{N}(v)$, which is determined by the local observation of agent $i$, i.e., the correspondence node $u$ of agent $j$ belongs to $\mathcal{N}(v)$ if and only if agent $j$ is in the observation range of agent $i$. The intuition behind this formulation is that agents are more likely to have an effect on each other when they can observe each other, especially when the multi-agent system itself is partially observable. The edge $e_{uv}$ between any two nodes in the graph $G$ is defined as:
\begin{equation}
    e_{uv}=\begin{cases}1,&\text{if}\ u \in \mathcal{N}(v)\ \text{or}\ v \in\mathcal{N}(u)\\0,&\text{otherwise}\end{cases}
\label{eq:adjmatrix}
\end{equation}
according to this definition, the adjacency matrix $A \in \mathbb{R}^{n\times n}$ is obtained. The feature matrix $X$ is constructed by combining the observation of each node, which is processed by the attention mechanism. The details of this attention mechanism will be illustrated in the next part. 

Then we build a three-layer graph convolutional network to learn the relation embedding vector of
each agent. The $l$-th graph convolutional layer is defined as:
\begin{equation}
H^{l+1}=\sigma\left(\tilde{D}^{-\frac{1}{2}} \tilde{A} \tilde{D}^{-\frac{1}{2}} H^{l} W^{l}\right),
\label{eq:gcn}
\end{equation}
where $\tilde{A} = A + I$,  $I$ is the identity matrix, $\tilde{D}$ is the degree matrix of $\tilde{A}$, $W^{l}$ is the parameter matrix that can be learned, $H^{l}$ is the input feature matrix of the graph convolutional layer and $H^{l+1}$ is the output of the graph convolutional layer, especially, $H^{0}=X$.

In practice, the output of the graph convolutional layer loses the original information of the input, resulting in the feature of the node being over-smooth. To solve this problem, we use a technique called skip connections to concatenate the original feature matrix with the $l$-th graph convolutional layer output as the input of the $(l+1)$-th graph convolutional layer, written as:
\begin{equation}
    H_{input}^{l+1}=CONCAT\left[H_{output}^{l}, X\right].
\end{equation}

QMIX restricts the parameters of the mixing network to be non-negative by using the absolute value of the parameters of the mixing network. Similarly, we use a softmax function on the output matrix $H^{3}$ of the three-layer graph convolutional network to enforce non-negativity, which we find empirically to have better performance than the standard absolute value function. With the weights being non-negative, we have ensured that the local optimal action is the same as the global optimal action, thus RACA satisfies the IGM condition:
\begin{equation}
    \frac{\partial Q_{\mathrm{tot}}(\boldsymbol{\tau}, \boldsymbol{u})}{\partial Q_{i}\left(\tau^{i}, u^{i}\right)} \geq 0, \quad \forall i \in \{1,\dots,n\},
\end{equation}
where $n$ represents the number of agents.

Then we use $softmax(H^{3})$ as the weight constraint on $ Q_{i}\left(\tau^{i}, u^{i}\right)$ to encode the topological structure between agents into the mixing network. RACA can be added to any mixing network such as QMIX because RACA does not change the structure of the mixing network. In this paper, we use the mixing network of QMIX as our mixing network.

\subsection{Observation Abstraction via Attention Mechanism}
To deal with the varying number of teammates during training and testing process, the algorithm needs to be population-invariant, we use an attention-based observation abstraction mechanism that can generalize to an arbitrary number of teammates with a fixed number of parameters.

The observation of agent $i$ is $o^{i}$, which is consisted of three parts: agent $i$'s own features $o_{own}^{i}$, population-variant features $o_{var}^{i}$, and population-invariant features $o_{inv}^{i}$. The size of population-variant features $o_{var}^{i}$ depends on the number of agents in agent $i$'s observation range. The size of population-invariant features $o_{inv}^{i}$ doesn't change over different maps. We first use a 1-layer fully connected network $h_i$ to process the input features and get the $Q$, $K$, $V$ matrix corresponding to $o_{own}^{i}$, $o_{var}^{i}$, $o_{var}^{i}$. Then we use scaled-dot product attention mechanism to generate an observation embedding using the $Q$, $K$, $V$ matrix as follows: 
\begin{equation}
\operatorname{Attention}(Q, K, V)=\operatorname{Softmax}\left(\frac{Q K^{T}}{\sqrt{d_{k}}}\right) V.
\end{equation}
Then we use a 1-layer fully connected network $g_i$ to process the embedding information generated by the local observation of the $i$-th agent.

Each agent has its own Q function and policy network. Particularly for agent $i$, its Q function is written as follows:
\begin{equation}
    Q_{i}\left(\tau^{i}, u^{i}\right)=f_{i}\left(g_{i}\left(\operatorname{Attention}(Q, K, V)\right), o_{inv}^{i}\right),
\end{equation}
where $f_{i}$ is a DRQN\cite{Hausknecht2015DeepRQ} network that takes the concatenation of the output of $g_i$ and the population-invariant features $o_{inv}^{i}$ of the $i$-th agent as input and outputs the final Q value $Q_{i}\left(\tau^{i}, u^{i}\right)$ of the $i$-th agent.

\subsection{Loss Function}
Our network is trained by minimizing the standard squared TD-error in DQN\cite{mnih2015human} to optimize our entire framework as follows:
\begin{equation}
    \mathcal{L}(\theta)=\sum_{i=1}^{b}\left[\left(y_{i}^{tot}-Q_{tot}(\boldsymbol{\tau},\boldsymbol{u}|\theta)\right)^{2}\right],
\end{equation}
where $y^{tot}=r+\gamma \max_{\boldsymbol{u^\prime}}Q_{tot}(\boldsymbol{\tau}^ \prime,\boldsymbol{u}^\prime|\theta^-)$ is the target joint action-value function. $b$ represents the batch size and $\theta^-$ are parameters of a periodically updated target network.

Our framework adopts the centralized training with decentralized execution paradigm. During centralized training, the learning algorithm use the mixing network to access the global state and the individual observation-action histories of all agents. During decentralized execution, agents choose actions based on their action-value function and can neither use the mixing network nor communicate with each other.

\begin{figure*}[ht]
    \centering
    \subfigure[2s3z]{
        \includegraphics[width=1.9 in]{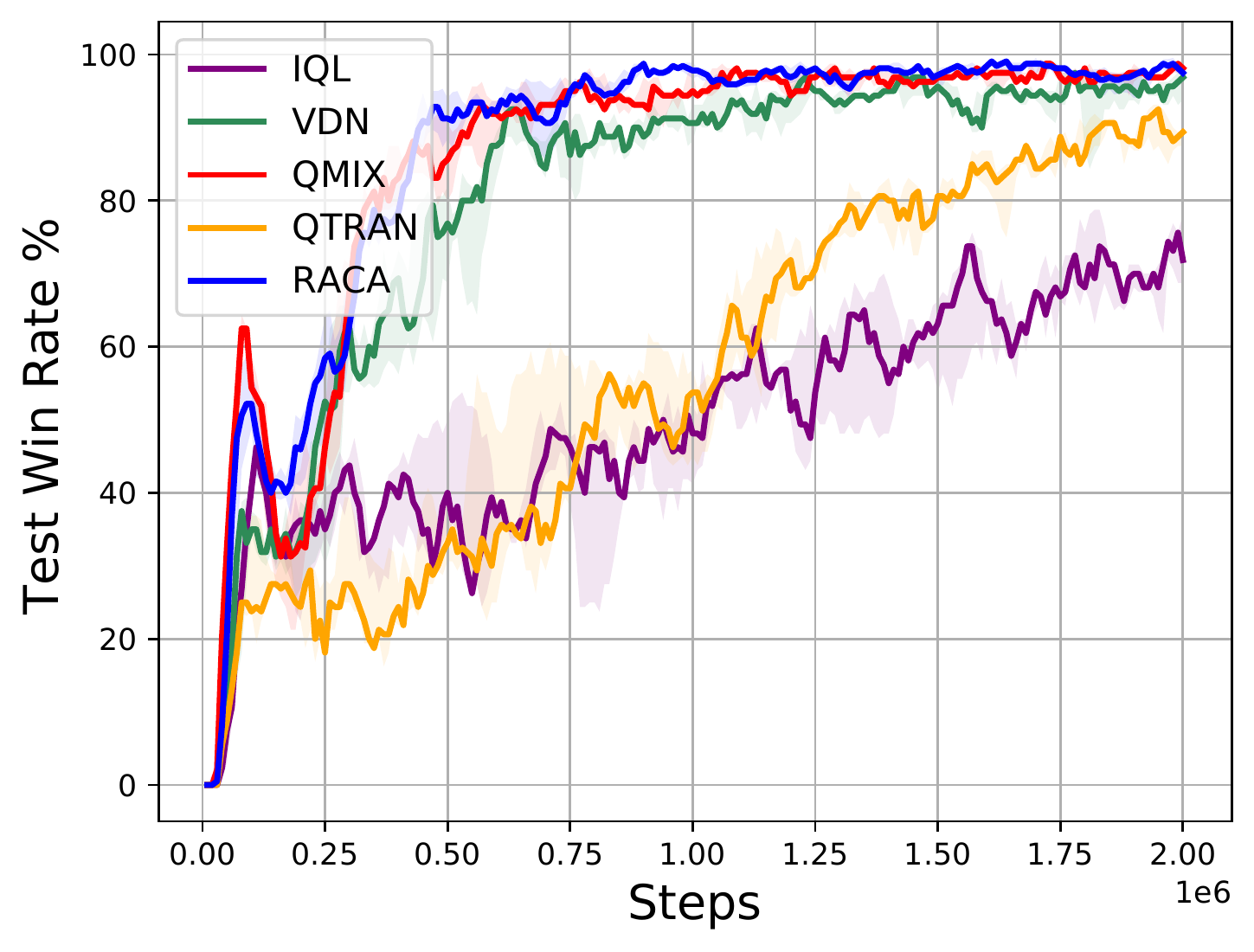}
    }
    \subfigure[3s5z]{
        \includegraphics[width=1.9 in]{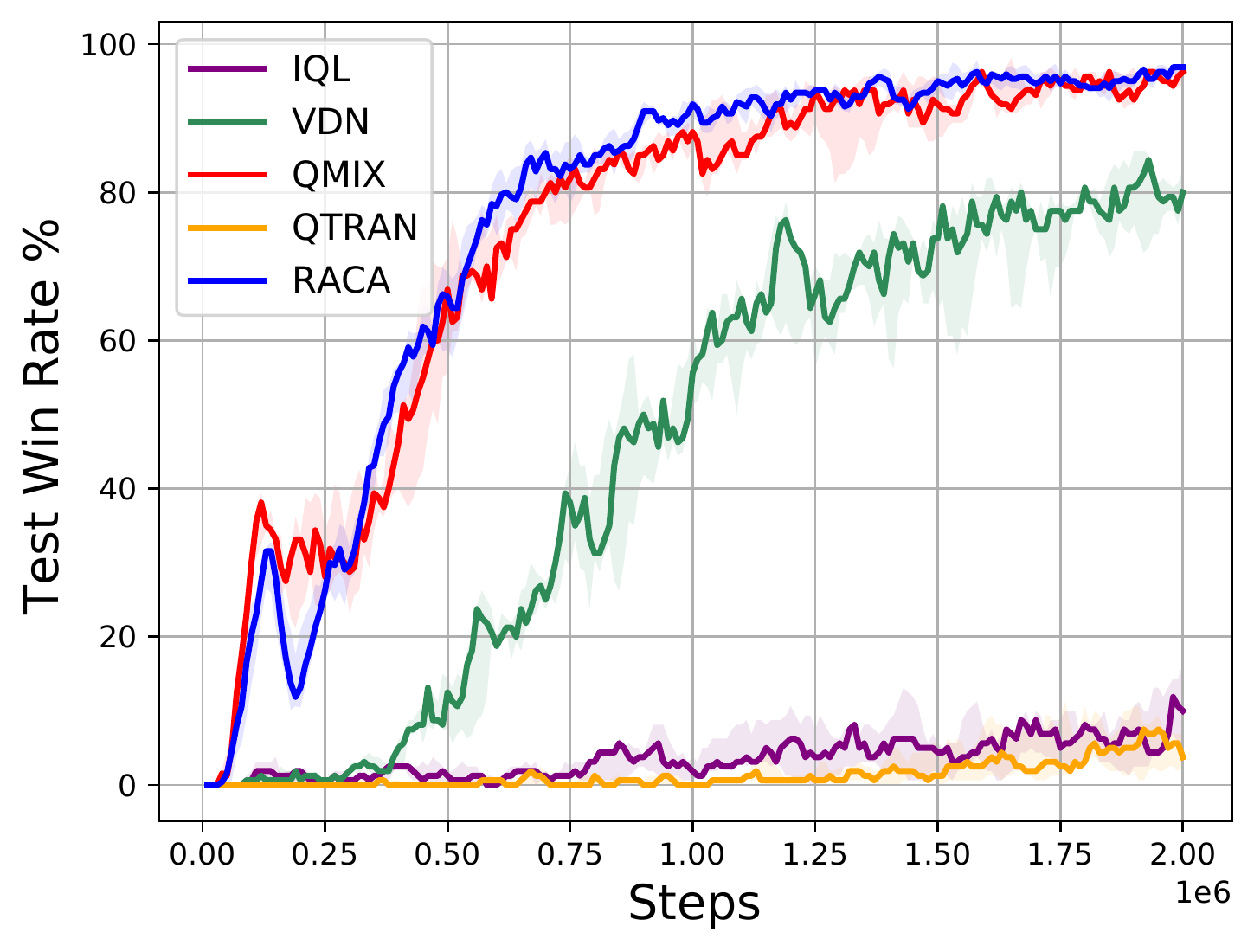}
    }
    \subfigure[1c3s5z]{
        \includegraphics[width=1.9 in]{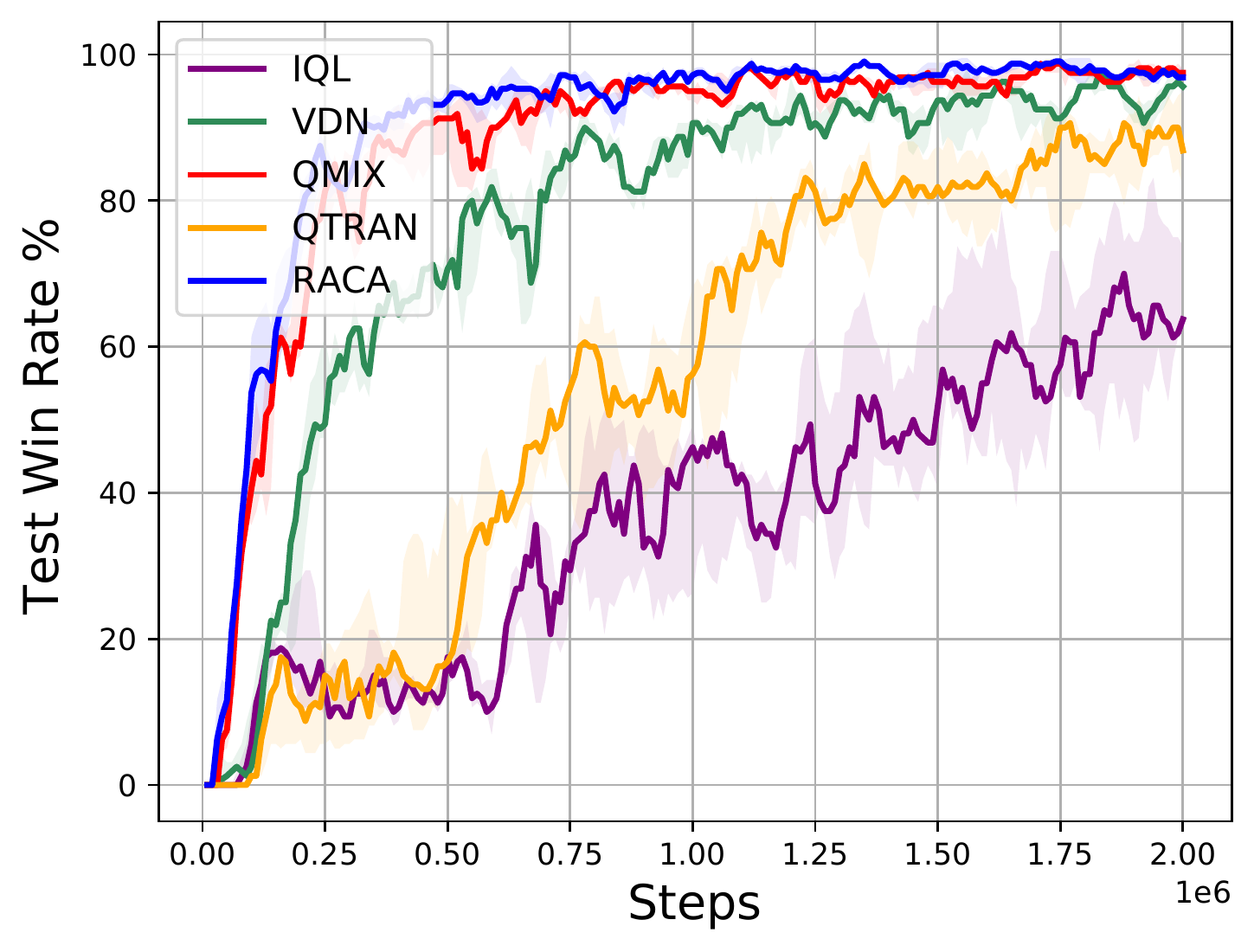}
    }

    \subfigure[MMM2]{
        \includegraphics[width=1.9 in]{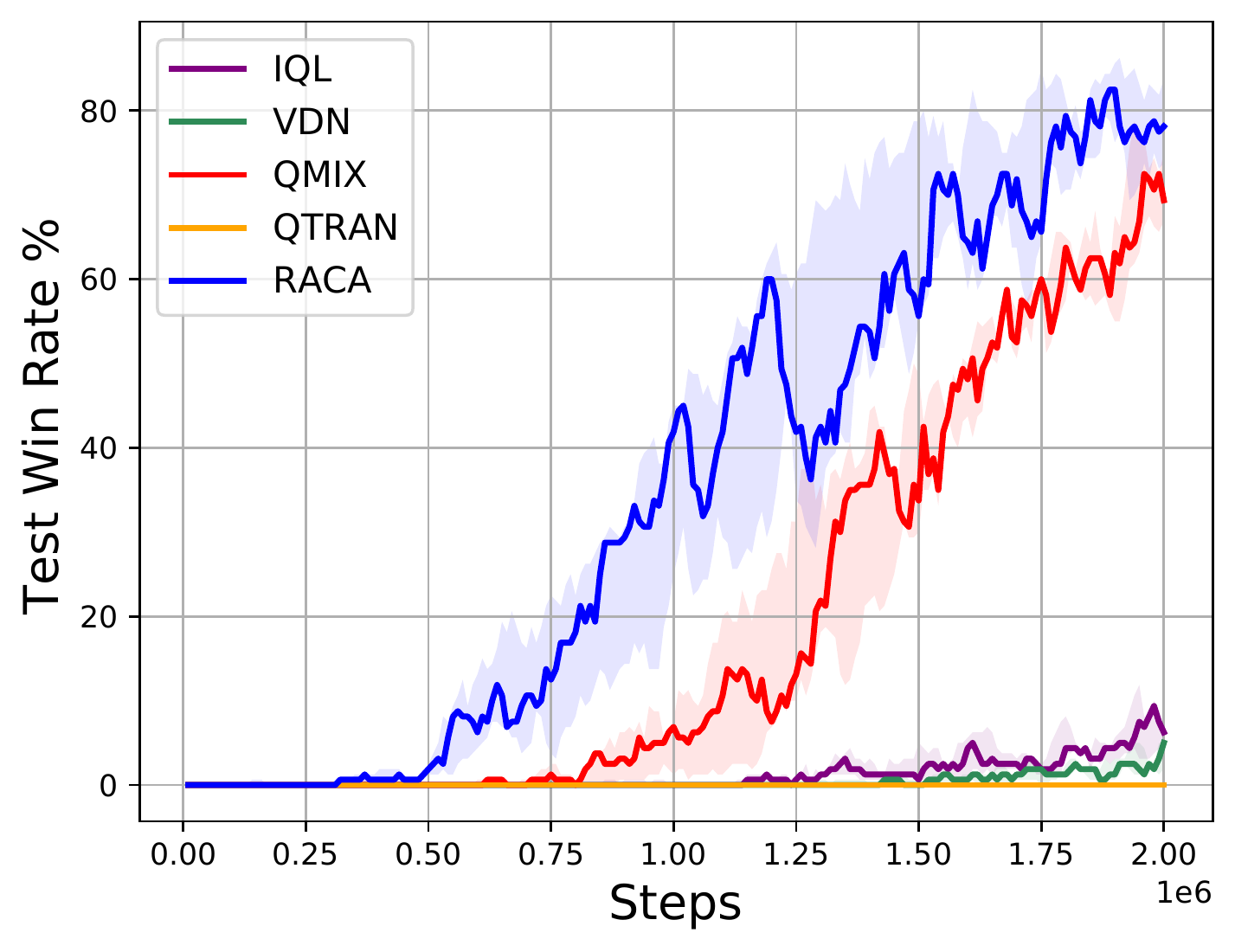}
    }
    \subfigure[5m\_vs\_6m]{
        \includegraphics[width=1.9 in]{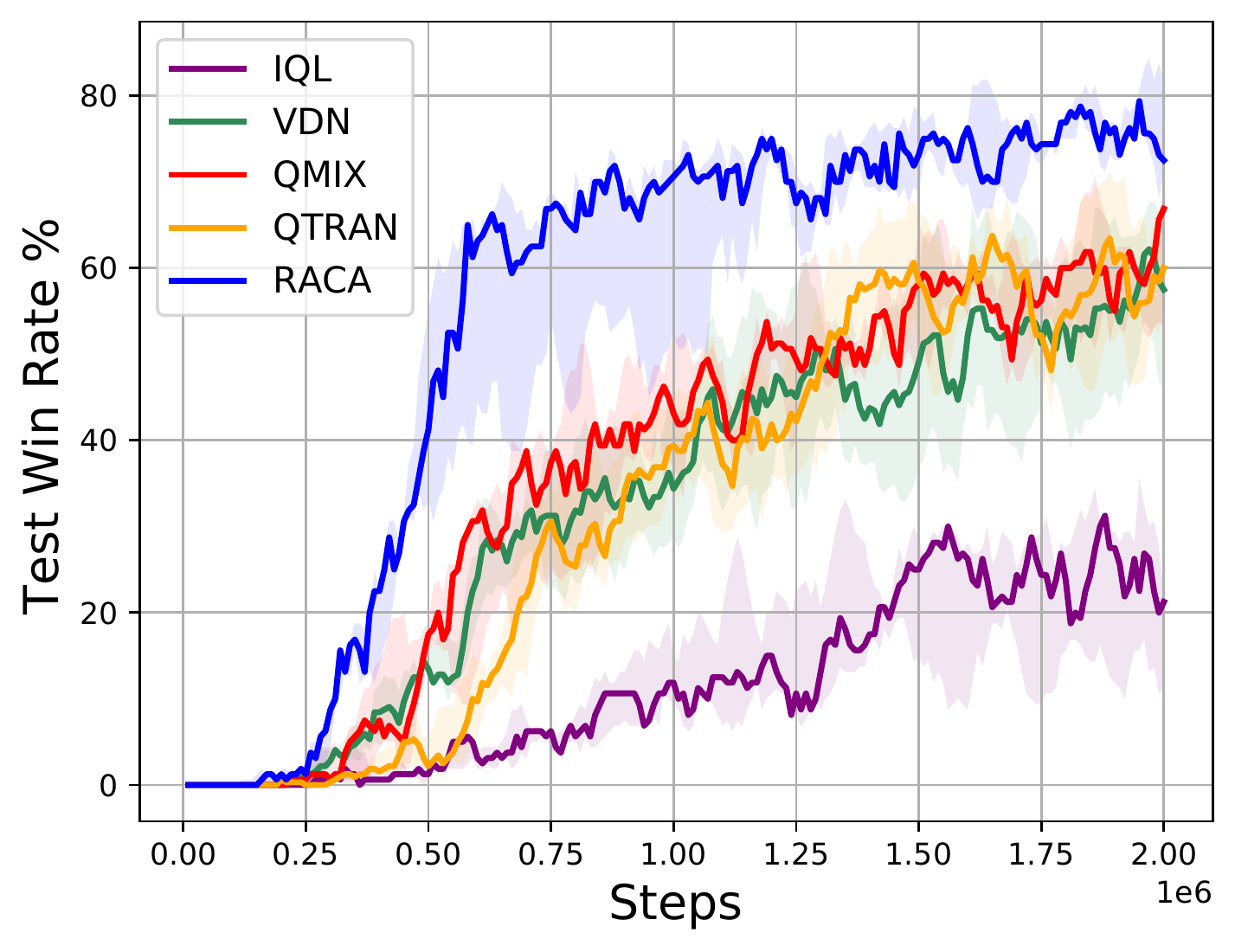}
    }
    \subfigure[8m\_vs\_9m]{
        \includegraphics[width=1.9 in]{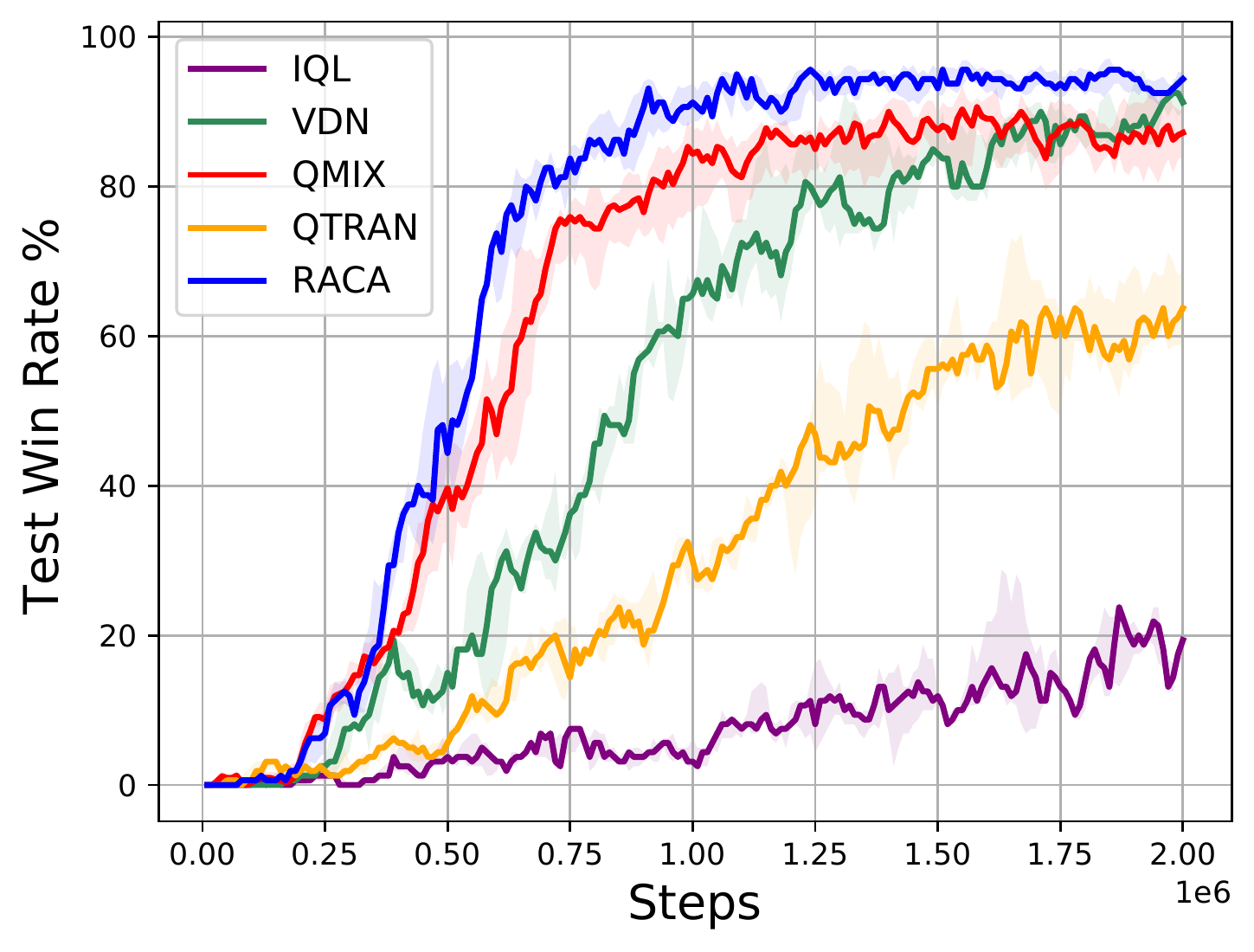}
    }
    \caption{Learning curves of our method and baseline algorithms on the StarCraftII micromanagement benchmark.}
    \label{fig:result1}
\end{figure*}

\section{Experiment}
In this section, we conduct experiments to answer the following questions: (1) Can our method improve learning efficiency? (2) Can our method promote zero-shot generalization to ad-hoc cooperation scenarios where agents change in number, category, or position? (3) To what extent does each component of our method contribute to the performance gains?

\subsection{Experiment Setup}

{\bf Environment}
We empirically evaluate our method on the StarCraft Multi-Agent Challenge (SMAC) benchmark for its high complexity of control and diversified environments. In the SMAC benchmark, agents learn in a discrete action space, including move (in four directions), stop, do nothing, and attack a certain enemy. SMAC provides easy, hard, and super hard scenarios according to the difficulty of the task. We choose representative scenarios of all three levels to conduct experiments. Multiple challenges are encountered in these scenarios including focus fire and macro tactics. Table\ref{table:scenario1} shows the detailed information of these scenarios. In addition, we conduct experiments on ad-hoc cooperation scenarios based on these above-mentioned representative scenarios. A detailed description of ad-hoc cooperation scenarios will be provided in the next section. The algorithms are trained by fighting with built-in game bots. For every 10,000 training steps, each algorithm is evaluated by running 32 testing episodes to get the current win rate and other related information.

{\bf Baselines and Ablations}
In this paper, we use independent Q-learning (IQL), and value function factorization MARL algorithms (VDN, QMIX, and QTRAN) as our baseline algorithms to compare with our proposed method. All of the baseline algorithms and ablation algorithms are listed in Table\ref{table:baselines}. For these baseline algorithms, we use the code provided in PyMARL\cite{samvelyan19smac}. For IQL\_Attn, VDN\_Attn, QMIX\_Attn, and QTRAN\_Attn, we use our implementation of scaled-dot product attention module on top of the original algorithm. In practice, algorithms need to be invariant to different unit types in StarCraftII. To achieve training across different unit types, we add an extra digit for terran units and zerg units. Thus each type of unit has a shield, which is set to zero for terran units and zerg units. For QMIX\_Gcn, we remove the attention-based observation abstraction module from RACA to verify its contribution. Note that QMIX\_Attn can also be seen as RACA without the graph-based relation encoder. 

{\bf Training and Testing}
Our method is implemented on the PyMARL framework and adopts the CTDE paradigm. During training, the global state information can be used by the mixing network. During testing, each agent must make decisions based on their local action-observation history and cannot communicate with each other. We use 5 random seeds to carry out experiments of each method on each map and demonstrates the mean test win rate, which represents the average percentage of winning episodes. The version of our StarcraftII is SC2.4.10(B75689). We carry out experiments on NVIDIA TITAN RTX GPU 24G.

\begin{table} [t]
    \caption{Baseline and ablation algorithms.}
    \label{table:baselines}
    \centering
    \begin{tabular}{crcrcr}
        \toprule
        \multicolumn{2}{c}{} &
        \multicolumn{2}{l}{Alg.} &
        \multicolumn{2}{l}{Description} \\
        \cmidrule(lr){1-2}
        \cmidrule(lr){3-4}
        \cmidrule(lr){5-6}
        
        \multicolumn{2}{c}{\multirow{5}{*}{\makecell{Related\\ Works}}} &  \multicolumn{2}{l}{IQL} & \multicolumn{2}{l}{Independent Q-learning} \\
        \multicolumn{2}{c}{} & \multicolumn{2}{l}{VDN} & \multicolumn{2}{l}{Additivity constraint} \\
        \multicolumn{2}{c}{} & \multicolumn{2}{l}{QMIX} & \multicolumn{2}{l}{Monotonicity constraint} \\
        \multicolumn{2}{c}{} & \multicolumn{2}{l}{QTRAN} & \multicolumn{2}{l}{Constraint-free} \\
        
        \cmidrule(lr){1-2}
        \cmidrule(lr){3-4}
        \cmidrule(lr){5-6}
        
        \multicolumn{2}{c}{\multirow{5}{*}{\makecell{Abla-\\tions}}} & \multicolumn{2}{l}{IQL\_Attn} & \multicolumn{2}{l}{IQL with attention} \\
        \multicolumn{2}{c}{} & \multicolumn{2}{l}{VDN\_Attn} & \multicolumn{2}{l}{VDN with attention} \\
        \multicolumn{2}{c}{} & \multicolumn{2}{l}{QMIX\_Attn} & \multicolumn{2}{l}{QMIX with attention} \\
        \multicolumn{2}{c}{} & \multicolumn{2}{l}{QMIX\_Gcn} & \multicolumn{2}{l}{\makecell[l]{QMIX with gcn module}} \\
        \multicolumn{2}{c}{} & \multicolumn{2}{l}{QTRAN\_Attn} & \multicolumn{2}{l}{\makecell[l]{QTRAN with attention}} \\
        \toprule
    \end{tabular}
\end{table}

\begin{figure*}[ht]
    \centering
    \subfigure[5m\_vs\_6m-6m\_vs\_6m]{
        \includegraphics[width=1.9 in]{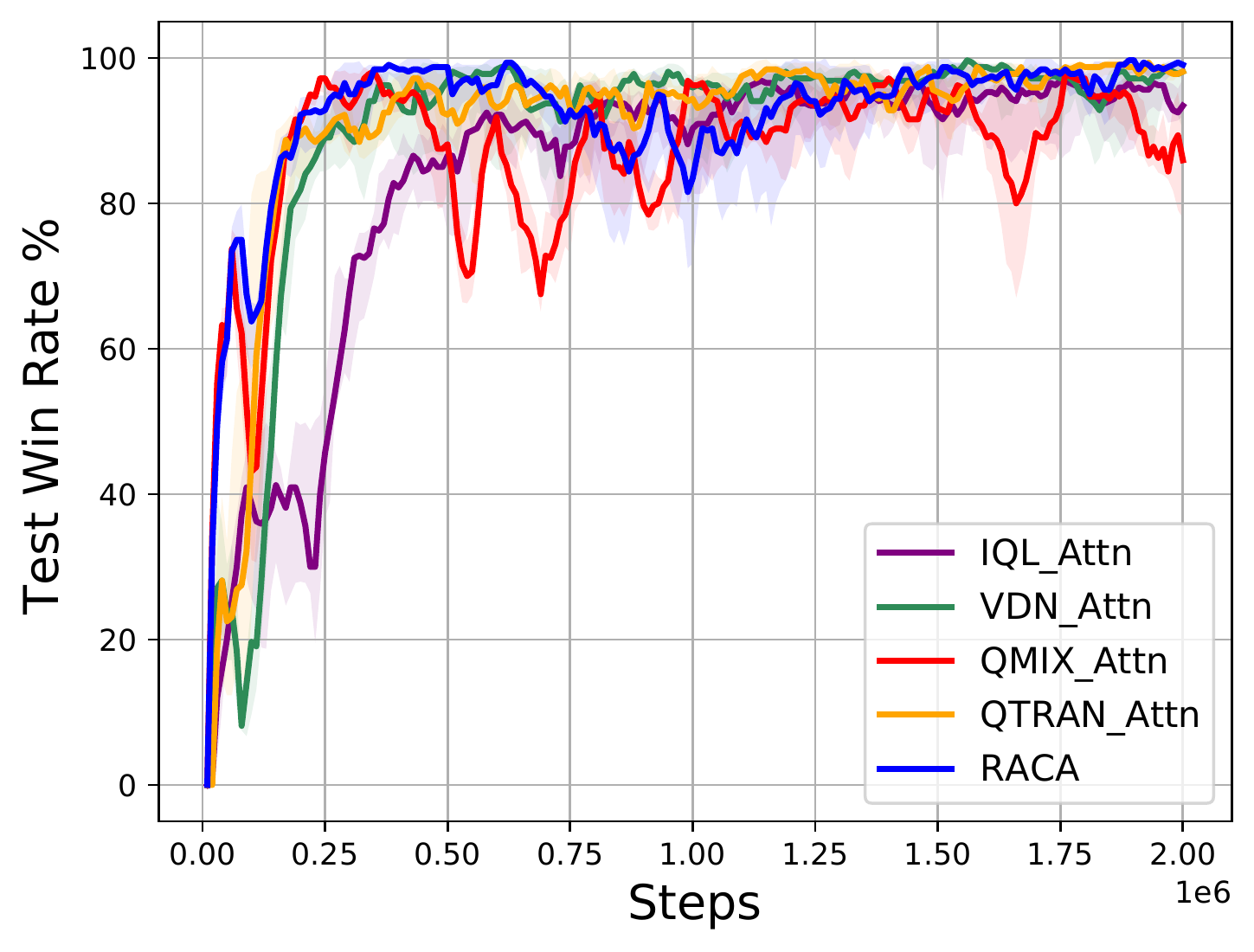}
    }
    \subfigure[3s2z-2s3z]{
        \includegraphics[width=1.9 in]{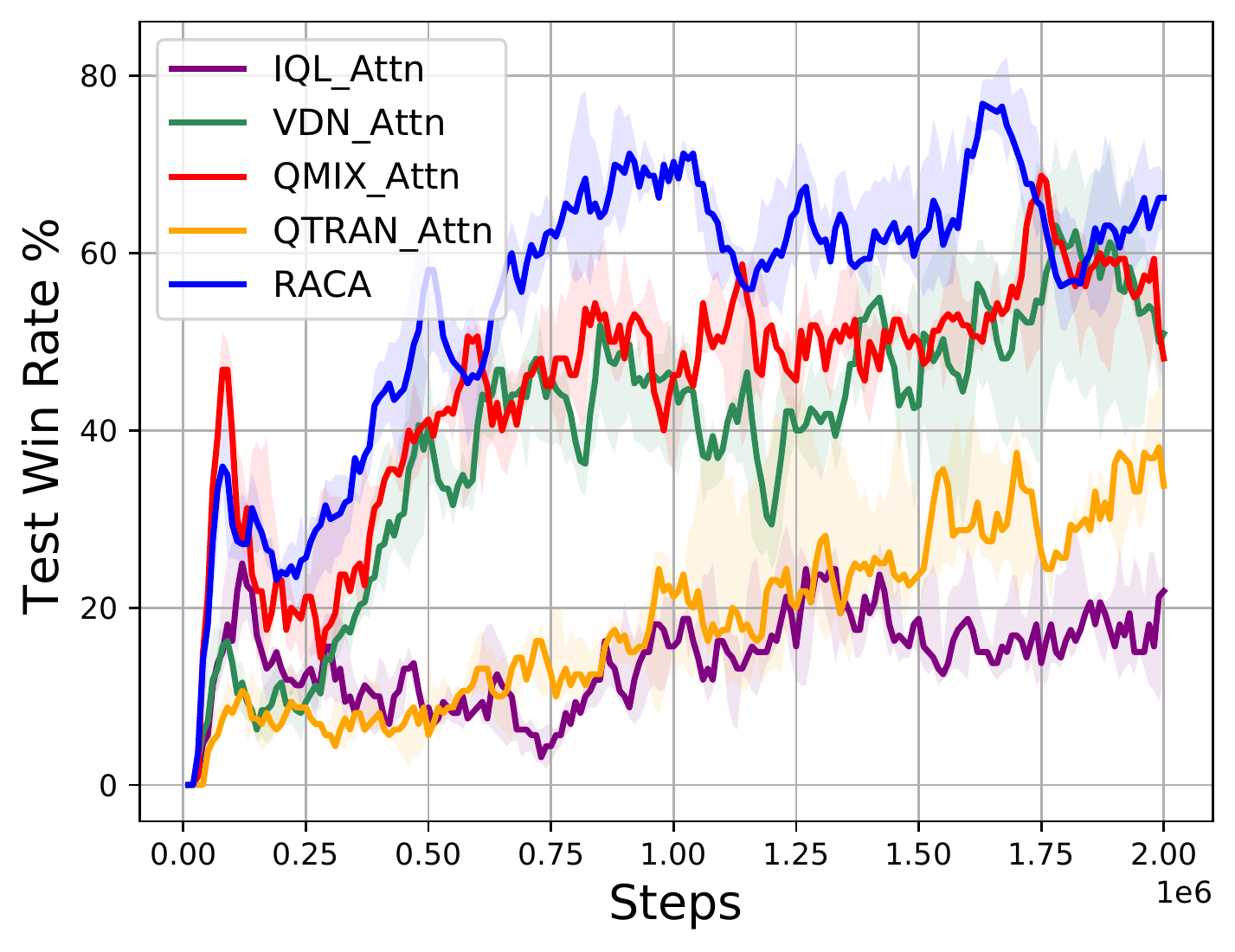}
    }
    \subfigure[2s3z-3s2z]{
        \includegraphics[width=1.9 in]{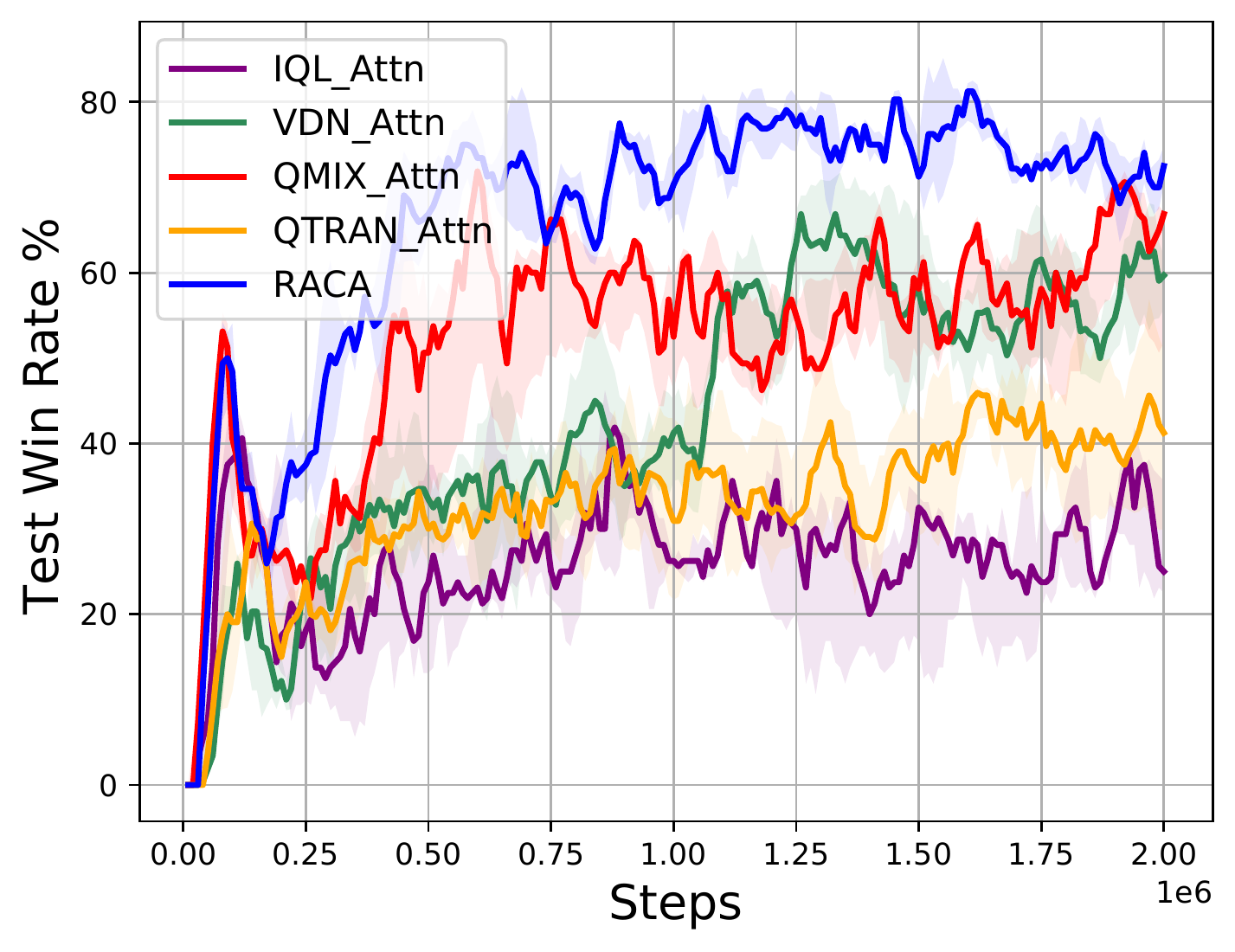}
    }

    \subfigure[5m-5ma]{
        \includegraphics[width=1.9 in]{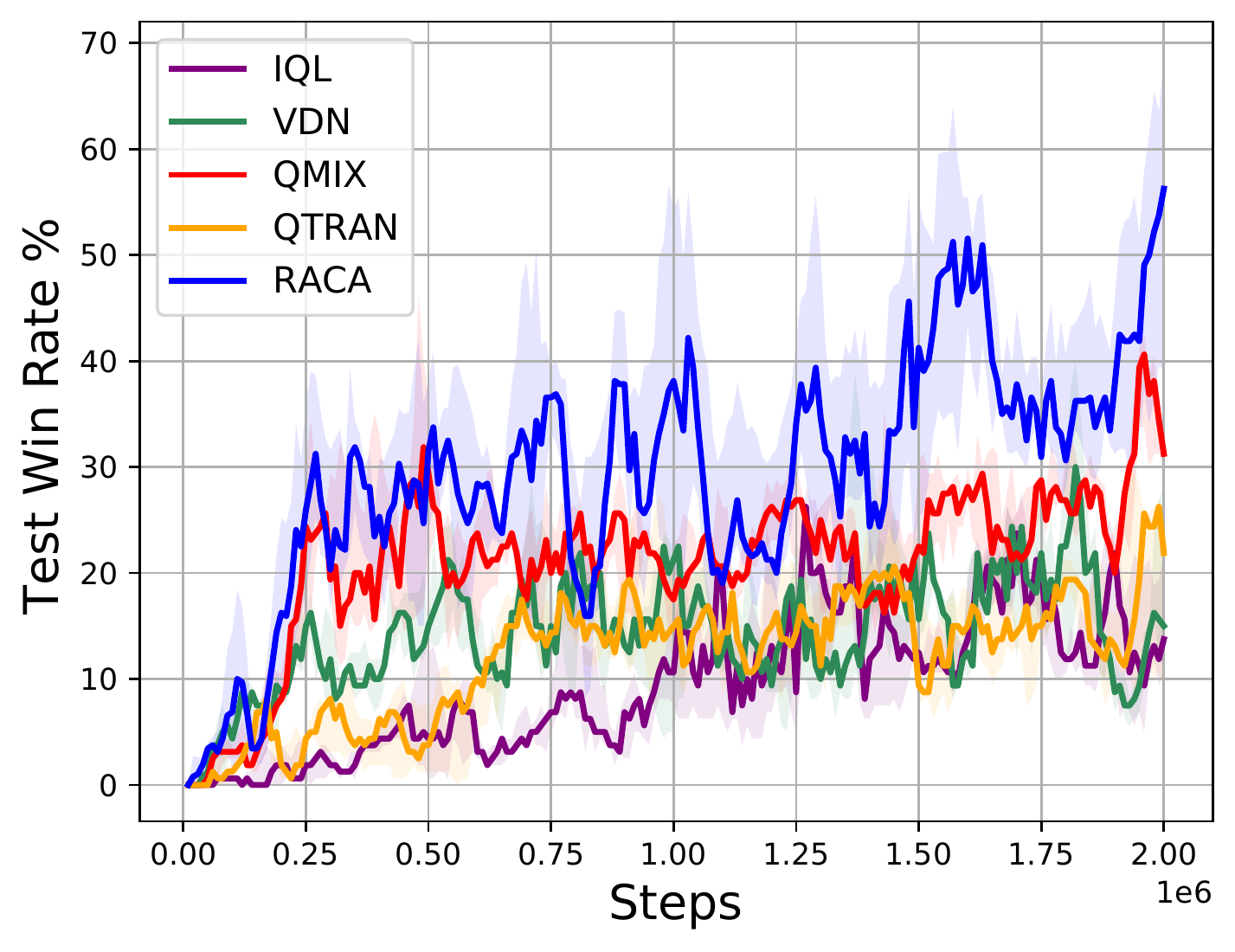}
    }
    \subfigure[5ma-5m]{
        \includegraphics[width=1.9 in]{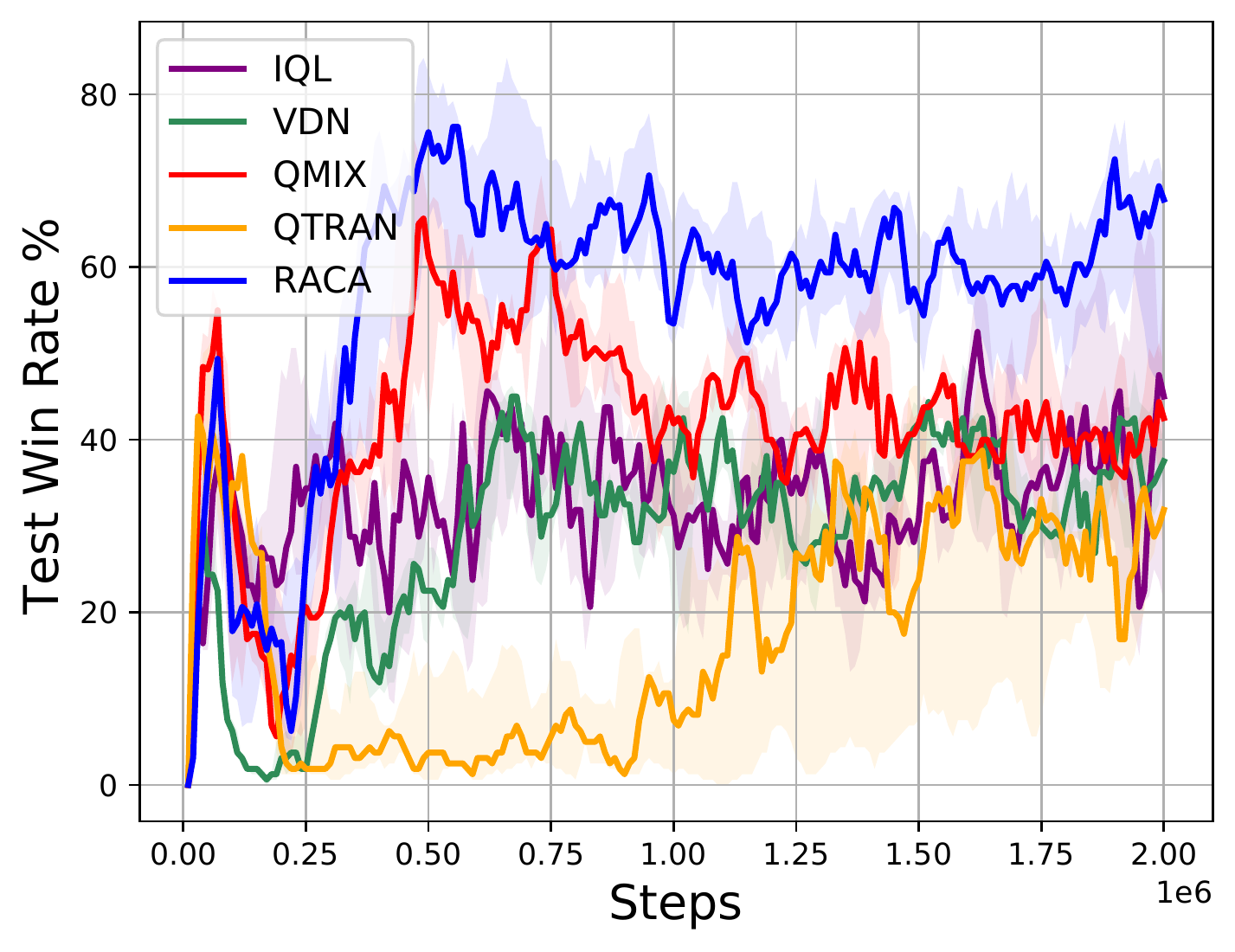}
    }
    \subfigure[MMM2\_rnd]{
        \includegraphics[width=1.9 in]{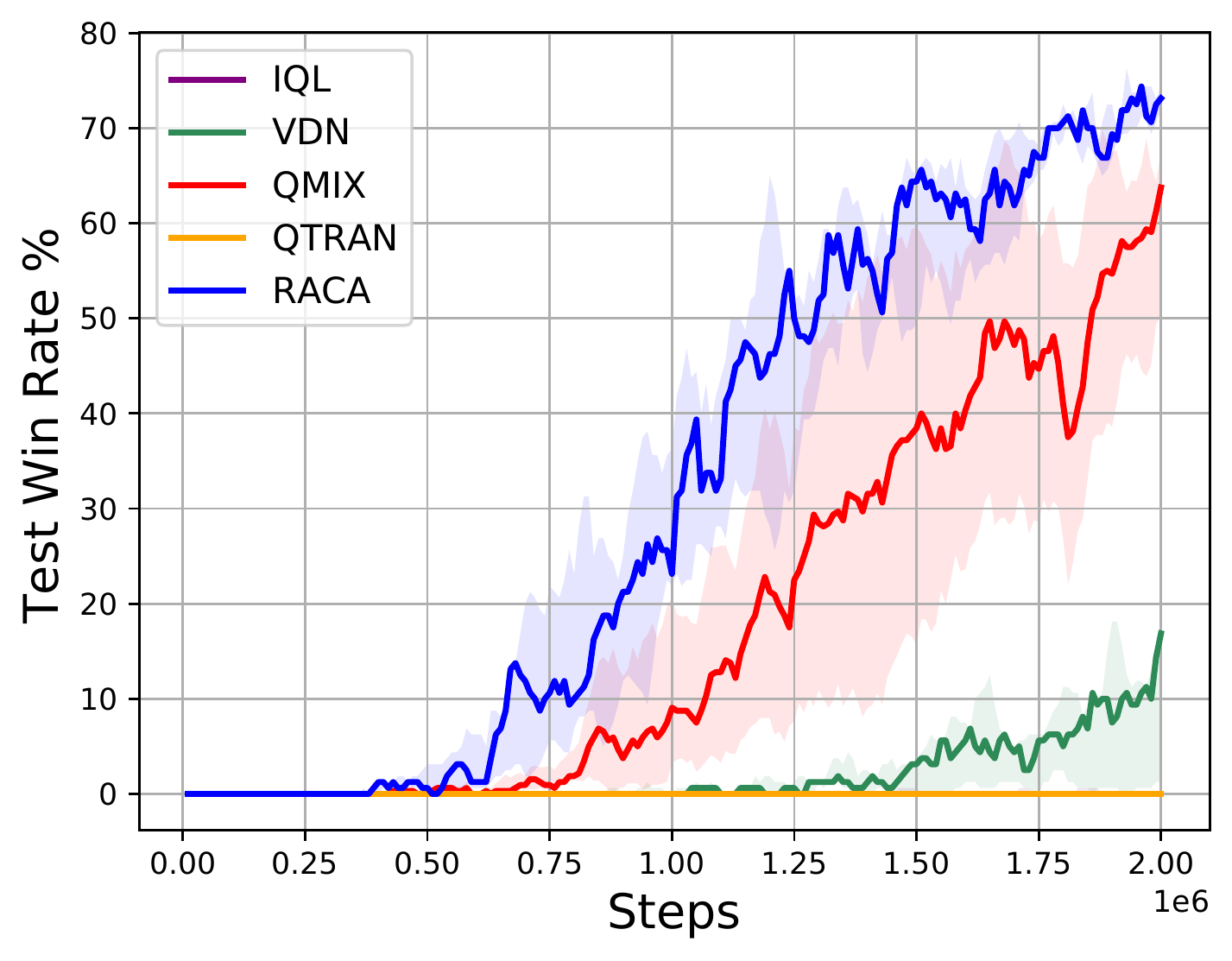}
    }
    \caption{Learning curves of our method and baseline algorithms on ad-hoc cooperation scenarios.}
    \label{fig:result2}
\end{figure*}

{\bf Hyperparameter Setting}
For all experiments, we use the same hyperparameters as the hyperparameters of QMIX in PyMARL and we use the default reward and observation settings of the SMAC benchmark. The whole framework is trained in an end-to-end fashion on fully unrolled episodes. Our proposed method is based on the mixing network of QMIX, which can be replaced by any mixing network such as the mixing network of VDN and QTRAN.

\subsection{Performance on the SMAC Benchmark}
To answer question (1), we evaluate our method on the SMAC benchmark. Figure\ref{fig:result1} demonstrates the superior performance of our proposed algorithm RACA on easy, hard, and super hard scenarios. Each solid line represents the mean test win rate and 25\%-75\% percentile is shaded. It can be seen that RACA has a better performance on all of the maps. Moreover, RACA outperforms baselines by a large margin on hard and super hard scenarios. However, RACA does not outperform baselines by a large margin on 3 easy maps. We suspect that this is because the graph-based relation encoder can make better use of the topological structure between agents when a complex strategy such as focus fire is needed.

\subsection{Performance on Ad-Hoc Cooperation Scenarios}
To answer question (2), we evaluate our method on ad-hoc cooperation scenarios. Each algorithm is trained and tested on different maps, for example, map 5m\_vs\_6m-6m\_vs\_6m means the algorithm is trained on map 5m\_vs\_6m and tested on map 6m\_vs\_6m, map MMM2\_rnd means the algorithm is trained and tested on map MMM2 with teammates at different positions. We propose four tasks to evaluate the ad-hoc cooperation performance of each algorithm: (a) cooperate with different number of teammates at test time (map 5m\_vs\_6m-6m\_vs\_6m) (b) cooperate with a different type of teammates at test time (map 5m-5ma and 5ma-5m) (c) cooperate with different number of teammates for each agent type at test time (map 2s3z-3s2z and 3s2z-2s3z) (d) cooperate with teammates at different positions at test time (map MMM2\_rnd).

Figure\ref{fig:result2} demonstrates the superior performance of our proposed algorithm RACA on ad-hoc cooperation scenarios. In some scenarios where the number of teammates is different at test time, we use baseline algorithms augmented with attention-based neural architectures for comparison. In other scenarios, we use the original baseline algorithms for comparison. Results show that RACA has a better performance on ad-hoc cooperation scenarios.

\begin{figure}[htp]
    \centering
    \includegraphics[width=1.9 in]{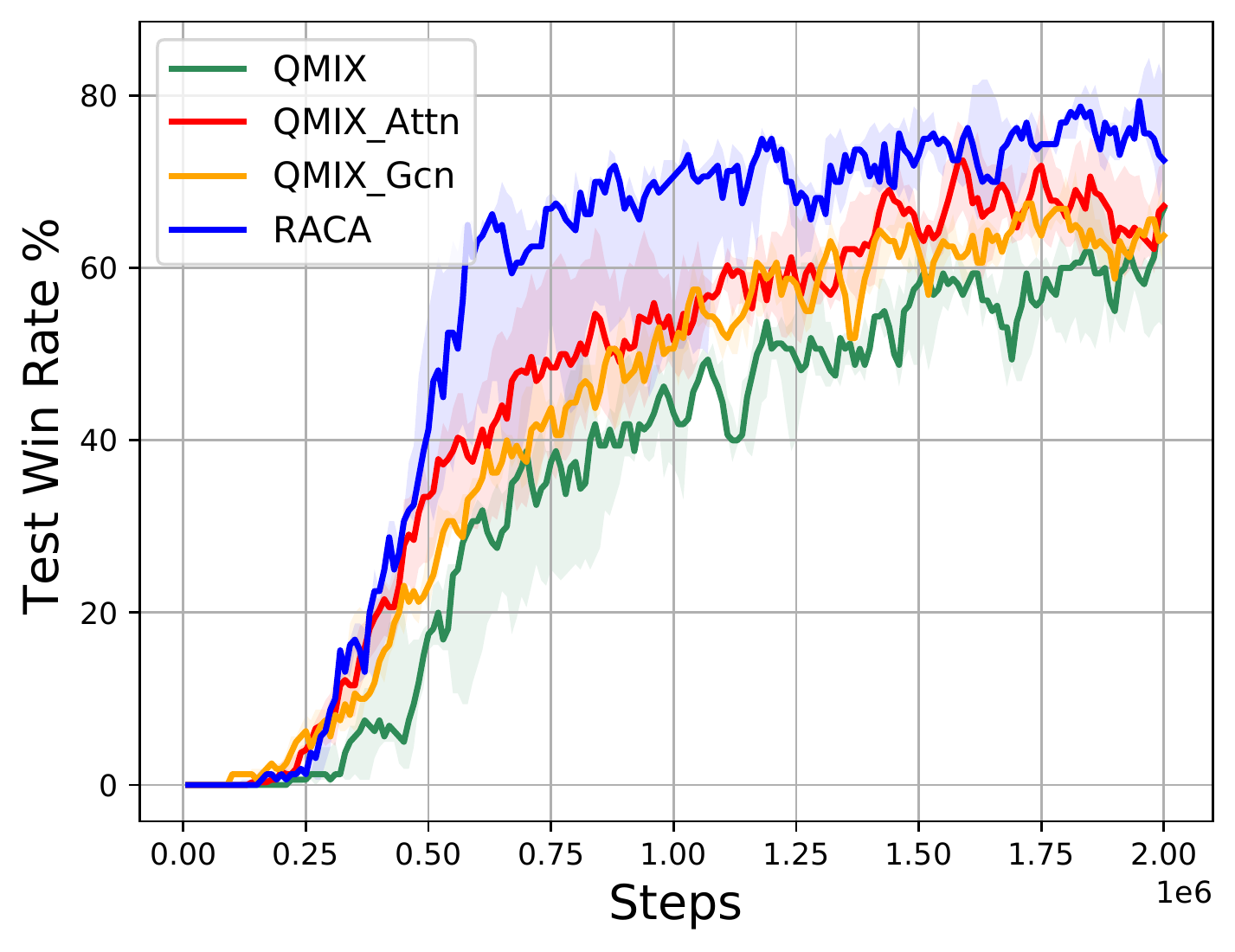}
    \caption{Ablation studies on map 5m\_vs\_6m.}
    \label{fig:ablation}
\end{figure}

\subsection{Ablation Studies}
To answer question (3), we carry out ablation studies to verify the contribution of each component of RACA. We remove the attention-based observation abstraction module from RACA and denote it as QMIX\_Gcn. To verify the contribution of the graph-based relation encoder, we remove it from RACA and denote it as QMIX\_Attn. As shown in figure~\ref{fig:ablation}, we compare RACA with QMIX\_Gcn, QMIX\_Attn, and QMIX on map 5m\_vs\_6m. Results show that the graph-based relation encoder and the attention-based observation abstraction module play a significant role in contributing to the performance gains. 

\section{Conclusion}
In this paper, we propose a novel multi-agent credit assignment method RACA, which utilizes the relationship between agents to achieve zero-shot generalization in ad-hoc cooperation scenarios. RACA takes advantage of a graph-based relation encoder to encode the topological structure between agents. Besides, RACA uses an attention-based observation abstraction mechanism that can generalize to an arbitrary number of teammates with a fixed number of parameters. Empirical results demonstrate that RACA outperforms baseline algorithms in terms of win rates on the StarCraftII micromanagement benchmark and ad-hoc cooperation scenarios. We believe the idea of utilizing the relationship between agents to promote ad-hoc cooperation can be an effective strategy for future works in multi-agent reinforcement learning.

\section{Acknowledgments}

This work is supported in part by the National Natural Science Foundation of China (Grand No. 61876181), Beijing Nova Program of Science and Technology under Grand No. Z191100001119043, the Youth Innovation Promotion Association, and CAS and the Projects of Chinese Academy of Science (Grant No. QYZDB-SSWJSC006).

\bibliographystyle{IEEEtran}
\bibliography{Reference}

\begin{thebibliography}{10}
\providecommand{\url}[1]{#1}
\csname url@samestyle\endcsname
\providecommand{\newblock}{\relax}
\providecommand{\bibinfo}[2]{#2}
\providecommand{\BIBentrySTDinterwordspacing}{\spaceskip=0pt\relax}
\providecommand{\BIBentryALTinterwordstretchfactor}{4}
\providecommand{\BIBentryALTinterwordspacing}{\spaceskip=\fontdimen2\font plus
\BIBentryALTinterwordstretchfactor\fontdimen3\font minus
  \fontdimen4\font\relax}
\providecommand{\BIBforeignlanguage}[2]{{%
\expandafter\ifx\csname l@#1\endcsname\relax
\typeout{** WARNING: IEEEtran.bst: No hyphenation pattern has been}%
\typeout{** loaded for the language `#1'. Using the pattern for}%
\typeout{** the default language instead.}%
\else
\language=\csname l@#1\endcsname
\fi
#2}}
\providecommand{\BIBdecl}{\relax}
\BIBdecl

\bibitem{bhalla2020deep}
S.~Bhalla, S.~Ganapathi~Subramanian, and M.~Crowley, ``Deep multi agent
  reinforcement learning for autonomous driving,'' in \emph{Canadian Conference
  on Artificial Intelligence}.\hskip 1em plus 0.5em minus 0.4em\relax Springer,
  2020, pp. 67--78.

\bibitem{ye2015multi}
D.~Ye, M.~Zhang, and Y.~Yang, ``A multi-agent framework for packet routing in
  wireless sensor networks,'' \emph{sensors}, vol.~15, no.~5, pp.
  10\,026--10\,047, 2015.

\bibitem{gao2021consensus}
Y.~Gao, W.~Wang, and N.~Yu, ``Consensus multi-agent reinforcement learning for
  volt-var control in power distribution networks,'' \emph{IEEE Transactions on
  Smart Grid}, vol.~12, no.~4, pp. 3594--3604, 2021.

\bibitem{yang2020multi}
Y.~Yang, Y.~Wen, J.~Wang, L.~Chen, K.~Shao, D.~Mguni, and W.~Zhang,
  ``Multi-agent determinantal q-learning,'' in \emph{International Conference
  on Machine Learning}.\hskip 1em plus 0.5em minus 0.4em\relax PMLR, 2020, pp.
  10\,757--10\,766.

\bibitem{foerster2018counterfactual}
J.~Foerster, G.~Farquhar, T.~Afouras, N.~Nardelli, and S.~Whiteson,
  ``Counterfactual multi-agent policy gradients,'' in \emph{Proceedings of the
  AAAI conference on artificial intelligence}, vol.~32, no.~1, 2018.

\bibitem{wang2020roma}
T.~Wang, H.~Dong, V.~Lesser, and C.~Zhang, ``Roma: Multi-agent reinforcement
  learning with emergent roles,'' \emph{arXiv preprint arXiv:2003.08039}, 2020.

\bibitem{Lowe2017MultiAgentAF}
R.~Lowe, Y.~Wu, A.~Tamar, J.~Harb, P.~Abbeel, and I.~Mordatch, ``Multi-agent
  actor-critic for mixed cooperative-competitive environments,'' in
  \emph{NIPS}, 2017.

\bibitem{rashid2018qmix}
T.~Rashid, M.~Samvelyan, C.~Schroeder, G.~Farquhar, J.~Foerster, and
  S.~Whiteson, ``Qmix: Monotonic value function factorisation for deep
  multi-agent reinforcement learning,'' in \emph{International Conference on
  Machine Learning}.\hskip 1em plus 0.5em minus 0.4em\relax PMLR, 2018, pp.
  4295--4304.

\bibitem{samvelyan19smac}
M.~Samvelyan, T.~Rashid, C.~S. Witt, G.~Farquhar, N.~Nardelli, T.~G.~J. Rudner,
  C.-M. Hung, P.~Torr, J.~N. Foerster, and S.~Whiteson, ``The starcraft
  multi-agent challenge,'' \emph{ArXiv}, vol. abs/1902.04043, 2019.

\bibitem{stone2010ad}
P.~Stone, G.~A. Kaminka, S.~Kraus, and J.~S. Rosenschein, ``Ad hoc autonomous
  agent teams: Collaboration without pre-coordination,'' in \emph{Twenty-Fourth
  AAAI Conference on Artificial Intelligence}, 2010.

\bibitem{Son2019QTRANLT}
K.~Son, D.~Kim, W.~Kang, D.~Hostallero, and Y.~Yi, ``Qtran: Learning to
  factorize with transformation for cooperative multi-agent reinforcement
  learning,'' \emph{ArXiv}, vol. abs/1905.05408, 2019.

\bibitem{wang2020qplex}
J.~Wang, Z.~Ren, T.~Liu, Y.~Yu, and C.~Zhang, ``Qplex: Duplex dueling
  multi-agent q-learning,'' \emph{arXiv preprint arXiv:2008.01062}, 2020.

\bibitem{xu2021mmd}
Z.~Xu, D.~Li, Y.~Bai, and G.~Fan, ``Mmd-mix: Value function factorisation with
  maximum mean discrepancy for cooperative multi-agent reinforcement
  learning,'' in \emph{2021 International Joint Conference on Neural Networks
  (IJCNN)}.\hskip 1em plus 0.5em minus 0.4em\relax IEEE, 2021, pp. 1--7.

\bibitem{liu2020multi}
Y.~Liu, W.~Wang, Y.~Hu, J.~Hao, X.~Chen, and Y.~Gao, ``Multi-agent game
  abstraction via graph attention neural network,'' in \emph{Proceedings of the
  AAAI Conference on Artificial Intelligence}, vol.~34, no.~05, 2020, pp.
  7211--7218.

\bibitem{jiang2018graph}
J.~Jiang, C.~Dun, T.~Huang, and Z.~Lu, ``Graph convolutional reinforcement
  learning,'' \emph{arXiv preprint arXiv:1810.09202}, 2018.

\bibitem{Oliehoek2016ACI}
F.~A. Oliehoek and C.~Amato, ``A concise introduction to decentralized
  pomdps,'' in \emph{SpringerBriefs in Intelligent Systems}, 2016.

\bibitem{sunehag2018value}
P.~Sunehag, G.~Lever, A.~Gruslys, W.~M. Czarnecki, V.~Zambaldi, M.~Jaderberg,
  M.~Lanctot, N.~Sonnerat, J.~Z. Leibo, K.~Tuyls \emph{et~al.},
  ``Value-decomposition networks for cooperative multi-agent learning based on
  team reward,'' in \emph{Proceedings of the 17th International Conference on
  Autonomous Agents and MultiAgent Systems}, 2018, pp. 2085--2087.

\bibitem{rashid2020weighted}
T.~Rashid, G.~Farquhar, B.~Peng, and S.~Whiteson, ``Weighted qmix: Expanding
  monotonic value function factorisation for deep multi-agent reinforcement
  learning,'' \emph{Advances in Neural Information Processing Systems},
  vol.~33, 2020.

\bibitem{guo2019attention}
S.~Guo, Y.~Lin, N.~Feng, C.~Song, and H.~Wan, ``Attention based
  spatial-temporal graph convolutional networks for traffic flow forecasting,''
  in \emph{Proceedings of the AAAI conference on artificial intelligence},
  vol.~33, no.~01, 2019, pp. 922--929.

\bibitem{eisen2020optimal}
M.~Eisen and A.~Ribeiro, ``Optimal wireless resource allocation with random
  edge graph neural networks,'' \emph{ieee transactions on signal processing},
  vol.~68, pp. 2977--2991, 2020.

\bibitem{fout2017protein}
A.~Fout, J.~Byrd, B.~Shariat, and A.~Ben-Hur, ``Protein interface prediction
  using graph convolutional networks,'' \emph{Advances in neural information
  processing systems}, vol.~30, 2017.

\bibitem{kipf2016semi}
T.~N. Kipf and M.~Welling, ``Semi-supervised classification with graph
  convolutional networks,'' \emph{arXiv preprint arXiv:1609.02907}, 2016.

\bibitem{Velickovic2018GraphAN}
P.~Velickovic, G.~Cucurull, A.~Casanova, A.~Romero, P.~Li{\`o}, and Y.~Bengio,
  ``Graph attention networks,'' \emph{ArXiv}, vol. abs/1710.10903, 2018.

\bibitem{hamilton2017inductive}
W.~Hamilton, Z.~Ying, and J.~Leskovec, ``Inductive representation learning on
  large graphs,'' \emph{Advances in neural information processing systems},
  vol.~30, 2017.

\bibitem{Xu2018RepresentationLO}
K.~Xu, C.~Li, Y.~Tian, T.~Sonobe, K.~Kawarabayashi, and S.~Jegelka,
  ``Representation learning on graphs with jumping knowledge networks,'' in
  \emph{ICML}, 2018.

\bibitem{su2020counterfactual}
J.~Su, S.~Adams, and P.~A. Beling, ``Counterfactual multi-agent reinforcement
  learning with graph convolution communication,'' \emph{arXiv preprint
  arXiv:2004.00470}, 2020.

\bibitem{liu2020pic}
I.-J. Liu, R.~A. Yeh, and A.~G. Schwing, ``Pic: permutation invariant critic
  for multi-agent deep reinforcement learning,'' in \emph{Conference on Robot
  Learning}.\hskip 1em plus 0.5em minus 0.4em\relax PMLR, 2020, pp. 590--602.

\bibitem{chen2020aateam}
S.~Chen, E.~Andrejczuk, Z.~Cao, and J.~Zhang, ``Aateam: Achieving the ad hoc
  teamwork by employing the attention mechanism,'' in \emph{Proceedings of the
  AAAI Conference on Artificial Intelligence}, vol.~34, no.~05, 2020, pp.
  7095--7102.

\bibitem{zhang2020multi}
T.~Zhang, H.~Xu, X.~Wang, Y.~Wu, K.~Keutzer, J.~E. Gonzalez, and Y.~Tian,
  ``Multi-agent collaboration via reward attribution decomposition,''
  \emph{arXiv preprint arXiv:2010.08531}, 2020.

\bibitem{mahajan2022generalization}
A.~Mahajan, M.~Samvelyan, T.~Gupta, B.~Ellis, M.~Sun, T.~Rockt{\"a}schel, and
  S.~Whiteson, ``Generalization in cooperative multi-agent systems,''
  \emph{arXiv preprint arXiv:2202.00104}, 2022.

\bibitem{bowling2005coordination}
M.~Bowling and P.~McCracken, ``Coordination and adaptation in impromptu
  teams,'' in \emph{AAAI}, vol.~5, 2005, pp. 53--58.

\bibitem{agmon2012leading}
N.~Agmon and P.~Stone, ``Leading ad hoc agents in joint action settings with
  multiple teammates.'' in \emph{AAMAS}, 2012, pp. 341--348.

\bibitem{stone2009leading}
P.~Stone, G.~A. Kaminka, and J.~S. Rosenschein, ``Leading a best-response
  teammate in an ad hoc team,'' in \emph{Agent-mediated electronic commerce.
  Designing trading strategies and mechanisms for electronic markets}.\hskip
  1em plus 0.5em minus 0.4em\relax Springer, 2009, pp. 132--146.

\bibitem{tambe1997towards}
M.~Tambe, ``Towards flexible teamwork,'' \emph{Journal of artificial
  intelligence research}, vol.~7, pp. 83--124, 1997.

\bibitem{grosz1996collaborative}
B.~Grosz and S.~Kraus, ``Collaborative plans for complex group action,''
  \emph{Artificial Intelligence}, 1996.

\bibitem{barrett2011empirical}
S.~Barrett, P.~Stone, and S.~Kraus, ``Empirical evaluation of ad hoc teamwork
  in the pursuit domain.'' in \emph{AAMAS}, 2011, pp. 567--574.

\bibitem{ijcai202166}
\BIBentryALTinterwordspacing
D.~Xing, Q.~Liu, Q.~Zheng, and G.~Pan, ``Learning with generated teammates to
  achieve type-free ad-hoc teamwork,'' in \emph{Proceedings of the Thirtieth
  International Joint Conference on Artificial Intelligence, {IJCAI-21}}, Z.-H.
  Zhou, Ed.\hskip 1em plus 0.5em minus 0.4em\relax International Joint
  Conferences on Artificial Intelligence Organization, 8 2021, pp. 472--478,
  main Track. [Online]. Available: \url{https://doi.org/10.24963/ijcai.2021/66}
\BIBentrySTDinterwordspacing

\bibitem{long2020evolutionary}
Q.~Long, Z.~Zhou, A.~Gupta, F.~Fang, Y.~Wu, and X.~Wang, ``Evolutionary
  population curriculum for scaling multi-agent reinforcement learning,''
  \emph{arXiv preprint arXiv:2003.10423}, 2020.

\bibitem{gu2021online}
P.~Gu, M.~Zhao, J.~Hao, and B.~An, ``Online ad hoc teamwork under partial
  observability,'' in \emph{International Conference on Learning
  Representations}, 2021.

\bibitem{li2019robust}
S.~Li, Y.~Wu, X.~Cui, H.~Dong, F.~Fang, and S.~Russell, ``Robust multi-agent
  reinforcement learning via minimax deep deterministic policy gradient,'' in
  \emph{Proceedings of the AAAI Conference on Artificial Intelligence},
  vol.~33, 2019, pp. 4213--4220.

\bibitem{rahman2021towards}
M.~A. Rahman, N.~Hopner, F.~Christianos, and S.~V. Albrecht, ``Towards open ad
  hoc teamwork using graph-based policy learning,'' in \emph{International
  Conference on Machine Learning}.\hskip 1em plus 0.5em minus 0.4em\relax PMLR,
  2021, pp. 8776--8786.

\bibitem{Hausknecht2015DeepRQ}
M.~J. Hausknecht and P.~Stone, ``Deep recurrent q-learning for partially
  observable mdps,'' in \emph{AAAI Fall Symposia}, 2015.

\bibitem{mnih2015human}
V.~Mnih, K.~Kavukcuoglu, D.~Silver, A.~A. Rusu, J.~Veness, M.~G. Bellemare,
  A.~Graves, M.~Riedmiller, A.~K. Fidjeland, G.~Ostrovski \emph{et~al.},
  ``Human-level control through deep reinforcement learning,'' \emph{nature},
  vol. 518, no. 7540, pp. 529--533, 2015.

\end{thebibliography}

\end{document}